\documentclass{article}

\usepackage{arxiv}

\usepackage{authblk} 

\usepackage[utf8]{inputenc} 
\usepackage[T1]{fontenc}    
\usepackage{hyperref}       
\usepackage{url}            
\usepackage{booktabs}       
\usepackage{amsfonts}       
\usepackage{nicefrac}       
\usepackage{microtype}      
\usepackage{lipsum}
\usepackage{graphicx}
\graphicspath{ {./images/} }

\usepackage{subcaption}
\usepackage{multirow}
\usepackage{soul}
\usepackage{xcolor}
\usepackage{float}
\sethlcolor{yellow}
\usepackage{amsmath}

\title{An Artificial Intelligence Framework for Measuring Human Spine Aging Using MRI}

\author[1]{Roozbeh Bazargani}
\author[1]{Saqib Abdullah Basar}
\author[1]{Daniel Daly-Grafstein}
\author[1]{Rodrigo Solis Pompa}
\author[1]{Soojin Lee}
\author[1]{Saurabh Garg}
\author[1]{Yuntong Ma}
\author[2, 3]{John A. Carrino}
\author[1,*]{Siavash Khallaghi}
\author[1,*]{Sam Hashemi}

\affil[1]{Prenuvo, Redwood City, CA, USA}
\affil[2]{Weill Cornell Medical College, New York, NY, USA}
\affil[3]{Hospital for Special Surgery, New York, NY, USA}
\affil[*]{Co-senior authors}

\usepackage{etoolbox}
\makeatletter
\patchcmd{\@maketitle}%
  {\vskip 0.4in \@minus 0.1in \center{\today}   \vskip 0.2in}%
  {\vspace{0.25in}}{}{}
\makeatother

\begin{document}
\maketitle
\begin{abstract}
The human spine is a complex structure composed of 33 vertebrae. It holds the body and is important for leading a healthy life. The spine is vulnerable to age-related degenerations that can be identified through magnetic resonance imaging (MRI). In this paper we propose a novel computer-vison-based deep learning method to estimate spine age using images from over 18,000 MRI series. Data are restricted to subjects with only age-related spine degeneration. Eligibility criteria are created by identifying common age-based clusters of degenerative spine conditions using uniform manifold approximation and projection (UMAP) and hierarchical density-based spatial clustering of applications with noise (HDBSCAN). Model selection is determined using a detailed ablation study on data size, loss, and the effect of different spine regions. We evaluate the clinical utility of our model by calculating the difference between actual spine age and model-predicted age, the spine age gap (SAG), and examining the association between these differences and spine degenerative conditions and lifestyle factors. We find that SAG is associated with conditions including disc bulges, disc osteophytes, spinal stenosis, and fractures, as well as lifestyle factors like smoking and physically demanding work, and thus may be a useful biomarker for measuring overall spine health.
\end{abstract}

\section{Introduction}
\label{sec:introduction}
The spine is a complex anatomical structure consisting of numerous vertebrae (nominally, 7 cervical, 12 thoracic, and 5 lumbar) and motion segments that serves as both a supportive axis and a protective encasement for the spinal cord. It is integral to biomechanical function and core stability, enabling a diverse range of movements, and plays a fundamental role in facilitating the transmission of neural signals by distributing the peripheral nervous system. Age-related changes and pathological alterations can substantially disrupt these functions, leading to reduced mobility, sensory impairments, loss of motor function, and decreased quality of life. The aging process of the spine is multifactorial, driven by an interplay of genetic predispositions, underlying pathologies, and lifestyle factors \cite{papadakis2011pathophysiology}. Radiological imaging is crucial for the identification of spinal conditions, which may involve various components, including intervertebral discs, facet (zygapophyseal) joints, neural canals, muscles, tendons, ligaments, and bony structures. Magnetic resonance imaging (MRI) provides superior visualization of the musculoskeletal system, enabling the assessment and quantification of age-related morphological changes. Consequently, there is an increasing interest in the development of automated methods for spine age estimation utilizing imaging to determine what a healthy spine typically looks like at different stages of life. The objective of this study was to explore the feasibility of predicting spine age using deep learning models applied to whole-spine MRI images.

\subsection{Spine Maturation and Degeneration}
Spine degeneration and stratification of spine patients have been of interest to radiologists.
Kim et al. \cite{kim1992mri} introduced the classification of lumbar disc herniation from MR images in 1992.
Pfirrmann et al. \cite{pfirrmann2001magnetic} developed a classification system for lumbar intervertebral disc degeneration based on MR images. They showed a T2-weighted MRI is reliable for grading disc degeneration.
Riesenburger et al. \cite{riesenburger2015novel} proposed a novel classification method for lumbar disc degeneration that incorporates endplate changes, the presence of a high-intensity zone, and a reduction in disc height. 
Gille et al. \cite{gille2014degenerative, gille2017new} proposed a classification system for degenerative spondylolisthesis of the lumbar spine that was consistent with age and health-related quality of life scales.

\subsection{Previous Work in Spine Degeneration Detection}
Previous work shows the feasibility of determining spine degeneration from MR images using deep learning.
Lu et al. \cite{lu2018deep} proposed a deep learning method to determine the severity of spinal canal stenosis, left foraminal stenosis, and right foraminal stenosis. The method consisted of 1) segmentation and labeling of vertebrae using U-Net; 2) extracting disc-level image volumes from sagittal and axial views; and 3) passing them through a Convolutional Neural Network (CNN) to determine the severity.
SpineOne \cite{he2021spineone} is a one-stage framework that detects lumbar vertebrae and discs and classifies them into normal and degenerative.
Hallinan et al. \cite{hallinan2021deep} proposed a deep learning method to automate the detection and grading of the central canal, lateral recess, and neural foraminal stenosis in lumbar spine MR images. They used a combination of faster Region-based Convolutional Neural Network (RCNN) to detect the region of interest and a CNN classifier to grade the condition.
Zheng et al. \cite{zheng2022deep} quantitated lumbar disc degeneration from MRI based on the feature extracted after segmenting vertebrae, discs, and the spinal cord and extracting feature points.
Yi et al. \cite{yi2023deep} used a combination of CNN and transformer to detect degenerative diseases in the cervical and lumbar regions from T2-weighted MR images taken in sagittal and axial views.
Chen et al. \cite{chen2024deep} proposed a two-stage framework using Mask RCNN. The disc and vertebrae were localized based on sagittal view and the degenerative conditions were classified based on sagittal and axial views.

These studies suggest that spine degeneration can be detected using deep learning approaches that mimic a visual inspection of the spine. Given that these conditions are correlated with aging, we hypothesize that whole spine MRI contains information that can be leveraged by deep learning models to predict spine age. 

\subsection{Previous Work in Age Estimation}
In the field of computer vision, most age estimation methods focused on two-dimensional images of the face \cite{shen2018deep, zhang2017age, shu2015personalized, kemelmacher2014illumination}. In medical imaging estimation of age based on brain MRI scans has been extensively researched~\cite{zhang2023age, cole2018brain, shah2024ordinal, shen2022attention, jonsson2019brain, lee2022deep, de2022mind, peng2021accurate, armanious2021age, cheng2021brain, zhang2022robust, cole2017predicting}. The estimation can be achieved through two approaches \cite{zhang2023age}:
\begin{itemize}
    \item Regression method: The model predicts a continuous value representing the chronological age. 
    \item Bins method: The model uses several bins and predicts the bin containing the chronological age, outputting the probability of the brain age belonging to each bin. The final age is computed as the expected value.
\end{itemize}

Various models and loss functions have been proposed to tackle this task.
Ordinal distance encoded regularization loss has been added to the cross entropy loss to train a CNN \cite{shah2024ordinal}.
Attention-guided deep learning was used to predict the gestational age from T2-weighted MRI \cite{shen2022attention}.
The Extreme Gradient Boosting (XGBoost) regression algorithm was applied to approximately 42,000 T1-weighted brain MRI series \cite{de2022mind}.
Peng et al. \cite{peng2021accurate} used a lightweight fully convolutional structure based on VGGNet \cite{simonyan2014very} architecture in T1-weighted structural MRI (sMRI).
Armanious et al. \cite{armanious2021age} proposed Age-Net, a deep CNN with a novel iterative data-cleaning algorithm to separate atypical-aging patients in T1-weighted MRI.
Two-Stage-Age-Network was proposed to improve the brain age estimation in T1-weighted MRI \cite{cheng2021brain}. The first-stage network estimates the brain age based on T1-weighted MRI and sex. Then, the age estimate is refined in the second-stage network. The inputs to the second stage are MRI image, sex, and the discretized brain age estimated by the first-stage network and the output is the residual that is added to the discretized estimated brain age.
Ensemble learning was applied to sMRI to improve the performance \cite{zhang2022robust}. Three separate ensemble frameworks were used for young, middle-aged, and older age groups to mitigate the dependency of model accuracy on age groups. Each framework contained four models: a support vector machine, a CNN, GoogleNet \cite{szegedy2015going}, and ResNet \cite{he2016deep}. In summary, these studies demonstrate the feasibility of assigning organ-specific biological age using deep learning methods on MR images. 

In the context of the spine, there has been research on the relationship between spine conditions, age, and occupation \cite{watanabe2006age, ruhli2005age, savage1997relationship}. Some studies \cite{khan2013neural, sneath2022objective} utilized machine learning techniques to estimate spine age from MR images. However, the performance was low with $R^2 = 0.28$ and a mean absolute error of 10.28 years for spine age compared to chronological age in Sneath et al. study \cite{sneath2022objective}. Moreover, these studies were limited by small datasets (60 and 70 MRI scans for training with test sets of 9 and 10 MRI scans, respectively) and relied on feature engineering rather than directly using the MRI scans as inputs. Furthermore, none of these studies utilized a deep learning approach and instead relied on classical methods such as random forests, extreme gradient boosting trees, and support vector machines.

\subsection{Contributions}
\begin{figure}[!t]
\centerline{\includegraphics[width=\textwidth]{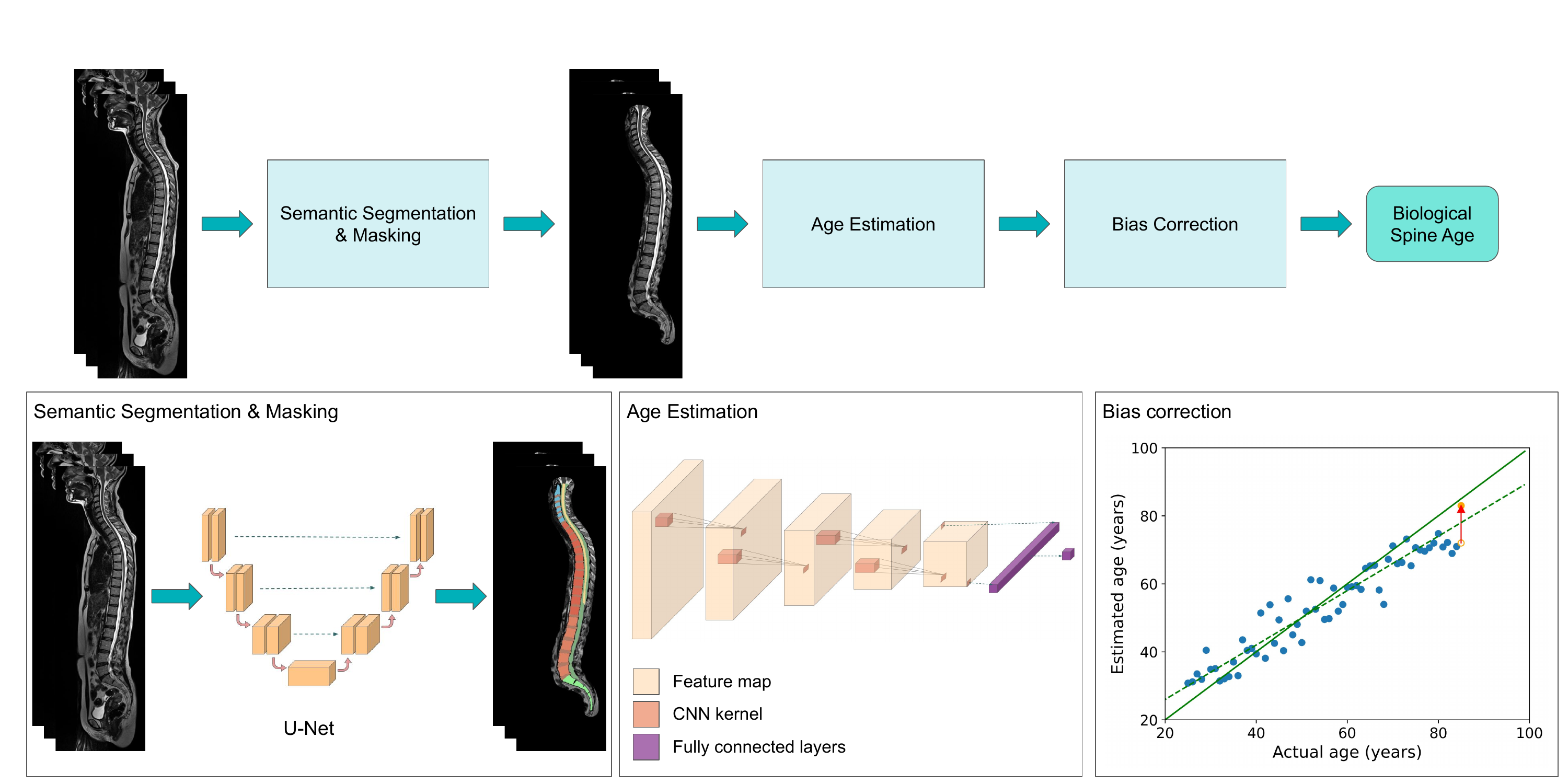}}
\caption{Overview of the three steps of the model to produce biological spine age. In the first step, semantic segmentation of the spine is generated and masked using a nnUnet model \cite{Khallaghi2023quantitative}. Next, the masked spine series are passed to a DCNN model consisting of multiple blocks of 3D convolution, batch-normalization, max-pooling layers, and a linear layer for the final prediction. The details of the model and layers are available in Table \ref{tab_model}. In the last step, we perform bias correction to produce bias-corrected spine age.}
\label{fig_model}
\end{figure}
We propose a Deep Convolutional Neural Network (DCNN) for predicting spine age based on a large dataset of sagittal T2-weighted MRI series.
\begin{itemize}
    \item To the best of our knowledge, this is the first work that utilizes MRI as input to a deep learning model for spine age estimation, achieving an $R^2$ performance of 0.87 compared to 0.28 in prior studies. This demonstrates that spine age can be estimated from T2-weighted MRI with high accuracy.
    \item We utilized an extensive dataset comprising of more than 18,000 series from more than 17,000 participants with over 8,500 test samples of normal and abnormal MR images. This significantly surpasses previous work in spine age estimation, which used only 70 series in total and 10 series in their test set.
    \item Our study demonstrates that the difference between chronological age and spine age, hereafter referred to as Spine Age Gap (SAG), is related to spinal conditions, pathologies, and lifestyle factors. This finding suggests that the SAG can serve as a valuable metric for assessing spine health.
\end{itemize}

\section{Material and Methods}
\subsection{Overview}
Figure~\ref{fig_model} shows an overview of our proposed spine age estimation model. The pipeline consists of three steps: 1) semantic segmentation and masking; 2) age estimation; and 3) bias correction. Next, we discuss the data and each of these steps in detail.

\subsection{Data}
We utilized a comprehensive dataset comprising of 18,070 3D T2-weighted whole spine MRI series. This dataset was acquired from 17,394 individuals with the sagittal view as the imaging plane. The scans were collected over a thirteen-year period, from 2011 to 2024, using 19 Philips and Siemens machines across 10 clinics in North America. The dataset included individuals aged 25 to 84 years, including those with supernumerary vertebrae (anomalous enumeration of vertebrae). Individuals were divided into six age groups, each spanning a 10-year interval, as shown in Table  \ref{tab_data}. The classification for 'normal' and 'abnormal,' as referenced in Table \ref{tab_data}, are explained in detail in the following section.


\begin{table}
\caption{Distribution of males and females in normal and abnormal series across age groups.}
\label{tab_data}
\setlength{\tabcolsep}{10pt}
\centering
\begin{tabular}{c|cc|cc}
\hline
Age bracket & \multicolumn{2}{|c|}{No. normal series} & \multicolumn{2}{|c}{No. abnormal series} \\
(years) & male & female & male & female \\
\hline
30 (25-34) & 561 & 521 & 404 & 411 \\
40 (35-44) & 1306 & 1238 & 1249 & 1252 \\
50 (45-54) & 1617 & 1379 & 857 & 949 \\
60 (55-64) & 1264 & 1299 & 645 & 683 \\
70 (65-74) & 598 & 622 & 377 & 378 \\
80 (75-84) & 122 & 84 & 132 & 122 \\
\hline
\multirow{2}{*}{Total} & 5468 & 5143 & 3664 & 3795 \\
& \multicolumn{2}{|c|}{10611} & \multicolumn{2}{|c}{7459} \\
\hline
\end{tabular}
\end{table}

\subsection{Establishing Normal Spines Based on Reports}\label{sec:methods_normal}
To train a DCNN model for predicting chronological age from MRI input, the dataset must be restricted to subjects who exhibit only age-related spine degeneration. To this end, we propose data-driven eligibility criteria to define participants who have a normal spine with respect to their age. 

We follow a clustering approach for identifying normal participants 
 based on spine conditions in their radiology report. This includes spinal structural and canal pathologies, as well as degenerative conditions for each vertebra. 
 
 The seven spinal structural and canal pathologies are: 1) bone lesion; 2) congenital spinal canal narrowing; 3) cord abnormalities; 4) fracture; 5) soft tissue edema; 6) spinal stenosis, and 7) spondylolisthesis. 
 
The eight degenerative spinal conditions consist of: 1) disc bulge; 2) disc osteophyte complex; 3) uncovertebral osteophyte; 4) protrusion; 5) extrusion; 6) desiccation; 7) endplate change;  and 8) annular fissure. In total, we looked at six cervical (C2-C7), thirteen thoracic (T1-T12, including T13 for cases with supernumerary vertebrae), and seven lumbar (L1-L5,  including L6-L7 for cases with supernumerary vertebrae) vertebrae that are $6+13+7=26$ vertebrae in total. Thus, we had $26\times8 + 7 = 215$ features for each report associated with each series.

This 215 feature vector is extremely sparse for the typical participant. Furthermore, radiologists frequently perform regional assessment of the spine and only report the most pertinent pathology in each region. We used this heuristic to convert this sparse feature vector into a dense representation. To this end, we aggregated affected vertebrae by region and severity of conditions. This reduced the number of features into $3\times8\times4=96$ for three regions (cervical, thoracic and lumbar) for eight degenerative spinal conditions and four severity values (mild, moderate, severe, and near complete). However, not all conditions are associated with all severity values listed above, which results in a smaller aggregated feature vector of size 60 for spinal conditions. By adding spinal structural and canal pathologies conditions we arrive at a final vector of size 67.   

In order to visualize this feature vector we performed a Uniform Manifold Approximation and Projection (UMAP)~\cite{mcinnes2018umap} dimensionality reduction to planar features for each age bracket. However, for our feature vector the default Euclidean distance used in UMAP evaluates different conditions and the frequency of conditions equally. In order to address this issue, we used the Canberra distance as defined below:
\begin{equation}d(p,q) = \sum_{i=1}^m \frac{|p_i - q_i|}{|p_i| + |q_i|},\end{equation}
where $m$ is the total number of conditions, $p$ and $q$ are vectors of the conditions. This formulation ensures that the distance between two cases where a specific condition has one more count is less than having a different condition. For instance, if participant 1 ($p1$) has 3 mild disc bulges, participant 2 ($p2$) has 4 mild disc bulges, and participant 3 ($p3$) has 3 mild disc bulges and 1 fracture, $d(p1, p2) = \frac{1}{7}$ and $d(p1, p3) = 1$.

Next, we applied Hierarchical Density-Based Spatial Clustering of Applications with Noise (HDBSCAN)~\cite{mcinnes2017hdbscan} to the reduced dimensions by UMAP to cluster samples. Within each age group clusters with more than 15\% of the population were considered normal, and clusters with less than 15\% were considered abnormal.

\subsection{Spine Age Estimation with Deep Convolutional Neural Network}
The field of view in a whole spine MRI spans multiple organs that age at different rates. In order to disentangle the spine from the rest of the organs, we first segment the spine using a semantic segmentation model~\cite{Khallaghi2023quantitative}. This process generates a segmentation mask that encompasses the cervical, thoracic, lumbar, and sacral vertebrae, intervertebral discs, ribs, cerebrospinal fluid, and the spinal cord. This mask is dilated and used to remove other regions from the MR image. In order to decrease the spatial variability of samples in our dataset, we resample all series to a common spacing of $0.9\times0.9\times3~mm^3$. Subsequently, we center cropped / padded all images to a fixed size of $384\times793\times14$.

\begin{table}
\caption{Layers of the model including the shape and number of parameters (weights + biases). Total number of parameters is 2,950,401.}
\label{tab_model}
\centering
\begin{tabular}{ll|l|r}
        \hline
        \multicolumn{2}{c|}{Layer} & \multicolumn{1}{c|}{Shape} & \multicolumn{1}{c}{Num Parameters} \\
        \hline
        \multirow{4}{*}{block1} & conv3d & [32, 1, 3, 3, 3] & 864 + 32 = 896 \\
        & batchnorm3d & [32] & 32 + 32 = 64 \\
        & relu & & \\
        & maxpool3d & & \\
        \hline
        \multirow{4}{*}{block2} & conv3d & [64, 32, 3, 3, 3] & 55296 + 64 = 55360 \\
        & batchnorm3d & [64] & 64 + 64 = 128 \\
        & relu & & \\
        & maxpool3d & & \\
        \hline
        \multirow{4}{*}{block3} & conv3d & [128, 64, 3, 3, 3] & 221184 + 128 = 221312 \\
        & batchnorm3d & [128] & 128 + 128 = 256 \\
        & relu & & \\
        & maxpool3d & & \\
        \hline
        \multirow{4}{*}{block4} & conv3d & [256, 128, 3, 3, 3] & 884736 + 256 = 884992 \\
        & batchnorm3d & [256] & 256 + 256 = 512 \\
        & relu & & \\
        & maxpool3d & & \\
        \hline
        \multirow{4}{*}{block5} & conv3d & [256, 256, 3, 3, 3] & 1769472 + 256 = 1769728 \\
        & batchnorm3d & [256] & 256 + 256 = 512 \\
        & relu & & \\
        & maxpool3d & & \\
        \hline
        \multirow{4}{*}{top} & conv3d & [64, 256, 1, 1, 1] & 16384 + 64 = 16448 \\
        & batchnorm3d & [64] & 64 + 64 = 128 \\
        & relu & & \\
        & maxpool3d & & \\
        \hline
        prediction & linear & [1, 64] & 64 + 1 = 65 \\
        \hline
    \end{tabular}
\end{table}

Our DCNN model was inspired by previous work in brain age estimation \cite{leonardsen2022deep, lee2024smoking}. The model is shown in Table~\ref{tab_model}. Mean Squared Error (MSE) loss was used to train the model.

After training, we used the validation data to correct the bias using Cole's method \cite{cole2018brain} as used in previous studies \cite{peng2021accurate, zhang2023age}. We computed the slope $\alpha$ and intercept $\beta$ of the fitted line to the samples using linear regression
\begin{equation}\hat{Y} = \alpha Y + \beta,\label{eq_cole1}\end{equation}
where $Y$ and $\hat{Y}$ represent chronological and predicted age, respectively. The bias-corrected predicted age $\hat{Y}_c$ is computed as
\begin{equation}\hat{Y}_c = \frac{\hat{Y} - \beta}{\alpha}.\label{eq_cole2}\end{equation}

\section{Experimental Settings}
\subsection{Normal Data}
\begin{figure}[!t]
    \centering
    \begin{subfigure}[b]{\textwidth}
        \centering
        \includegraphics[width=\textwidth]{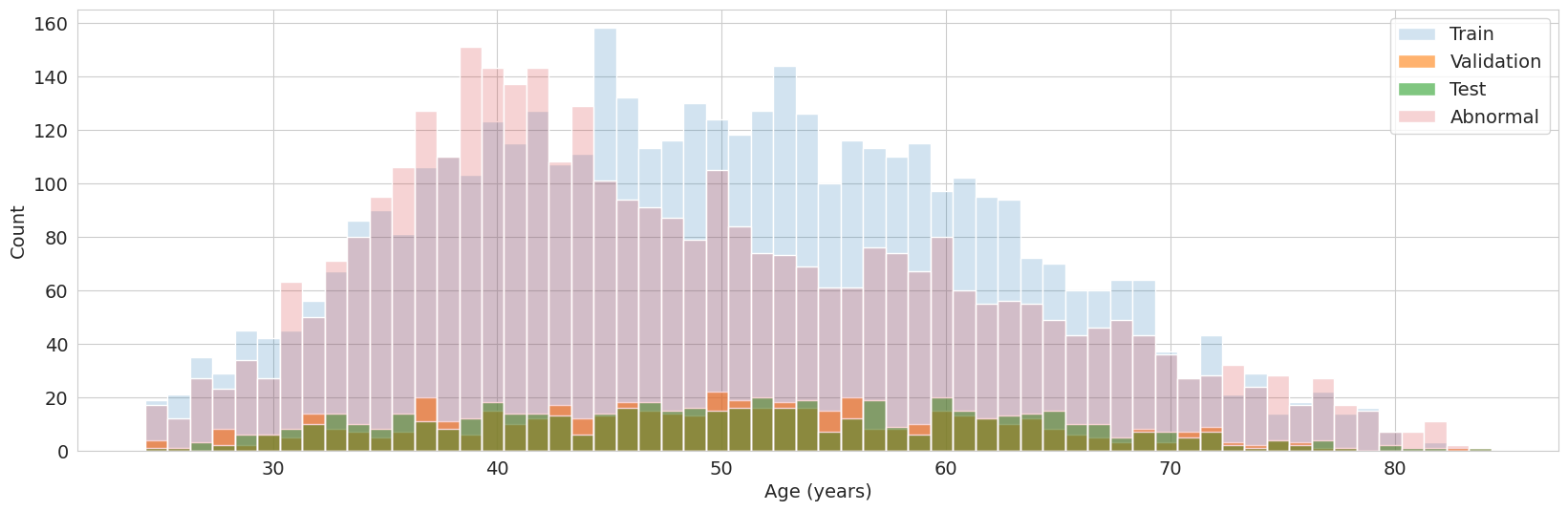}
        \caption{Male participants.}
    \end{subfigure}
    \begin{subfigure}[b]{\textwidth}
        \centering
        \includegraphics[width=\textwidth]{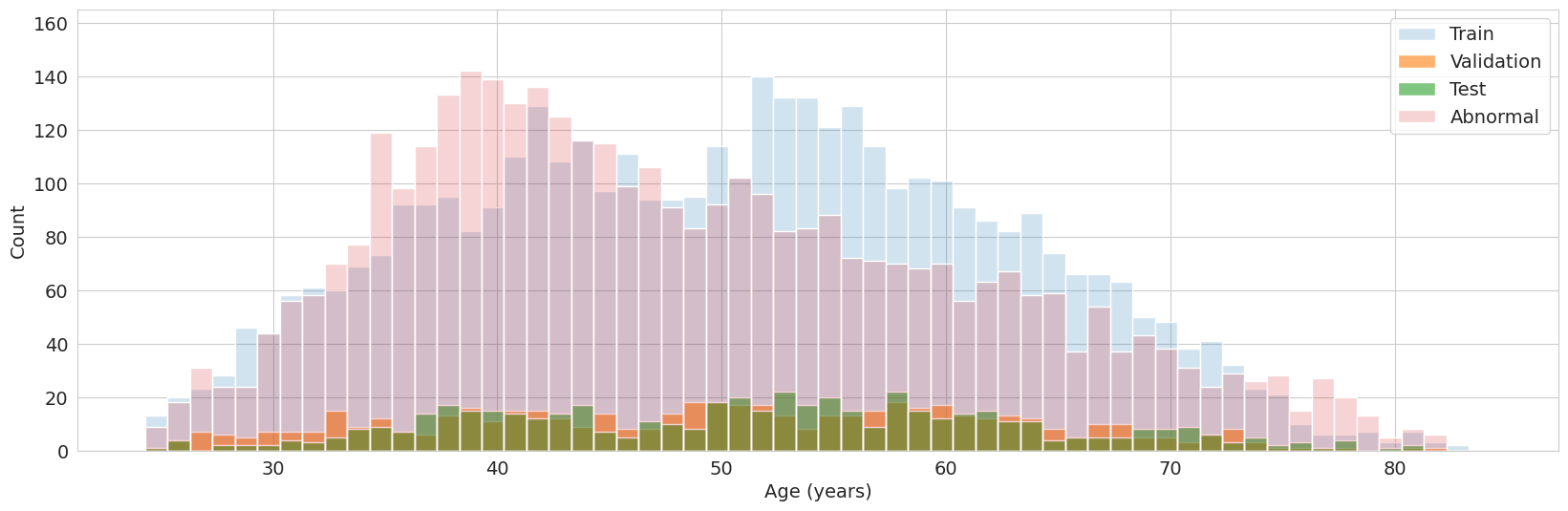}
        \caption{Female participants.}
    \end{subfigure}
    \caption{Train, validation, and test set histogram based on gender across age brackets.}
    \label{fig_split}
\end{figure}

We only used the images from individuals clustered as normal (see Section~\ref{sec:normal}) within their age group for training our DCNN model. This resulted in 10,611 series for the development of the spine age estimation model. We divided the dataset into training, validation, and test sets with 8,491, 1,051, and 1,069 samples, respectively. The samples were split while maintaining the gender and age ratios, as shown in Figure~\ref{fig_split}. To evaluate the relationship of spine conditions and lifestyle with predicted spine age, we mixed the abnormal data with the test set, resulting in a set of $7459 + 1069 = 8528$ samples for the full-test set.

\subsection{Sample size, Loss, and Spine regions}
To assess the proposed model, we compared it with models trained on a smaller number of samples, different loss, and different regions of the spine. In particular, we trained the same model on 85 (85 train and 10 validation) and 850 (850 train and 100 validation) samples. In addition, we trained the model using smooth-L1 loss. Similar to L1 loss, smooth-L1 loss helps mitigate the sensitivity to noisy series. We also trained the model on masked series that only included cervical, thoracic, and lumbar regions. Having segmentations of cervical, thoracic, and lumbar vertebrae, we masked those regions after dilation and trained models on the masked series.

The metrics used for comparison are mean absolute error (MAE), age bracket weighted MAE (WMAE), and $R^2$ between the actual and predicted age. A lower absolute error indicates better performance, as it shows that the model is predicting age more accurately. MAE is computed as:
\begin{equation}\text{MAE} = \frac{1}{n} \sum\limits_{i=1}^n |y_i - \hat{y}_i|,\end{equation}
where $y_i$ and $\hat{y}_i$ are $i$-th participant's chronological and predicted age and $n$ is the number of participants.

WMAE is calculating by computing the MAE in each age category, then taking the global mean of these age category errors as:
\begin{equation}\text{WMAE} = \frac{1}{N} \sum\limits_{j=1}^{N} (\frac{1}{n_j} \sum\limits_{i=1}^{n_j} |y_i - \hat{y}_i|),\end{equation}
where $N$ is the number of age brackets and $n_j$ is the number of samples in each age bracket. This allows all age categories to contribute equally in model evaluation regardless of the number of samples. 

$R^2$ has a range of $-\infty$ to $1$ and the higher values show better performance. It is computed as
\begin{equation}R^2 = 1 - \frac{\sum\limits_{i=1}^{n} (y_i - \hat{y}_i)^2}{\sum\limits_{i=1}^{n} (y_i - \bar{y})^2},\end{equation}
where $\bar{y}$ is the mean of participants' age. 

\subsection{Hyperparameters}
\subsubsection{UMAP}
We set the number of neighbors to fifteen, and the minimum distance to zero.

\subsubsection{HDBSCAN}
We set the minimum number of samples in a cluster to $1\%$ of the population in each age bracket. If a cluster has a smaller number of samples, it is considered an outlier. We set the number of samples in a neighborhood for a sample to be considered as a core sample to 5. Finally, we set the distance threshold that merges the clusters whose distance is below this threshold to 0.3 for the 70 and 80 age brackets, 0.7 for 40 and 60, and 1 for 30 and 50 based on the distribution of the sample points in the UMAP plot.

\subsubsection{DCNN}
For training the model, the batch size was set to two. Adam's optimizer was used with a learning rate of $0.01$. We also used a reduce learning rate on plateau scheduler with a factor of $0.3$ and a patience of five.

\section{Results and Discussion}
\subsection{Establishing the Normal Spine}~\label{sec:normal}
\begin{figure}[!t]
    \centering
    \begin{subfigure}[b]{0.47\textwidth}
        \centering
        \includegraphics[width=0.82\textwidth]{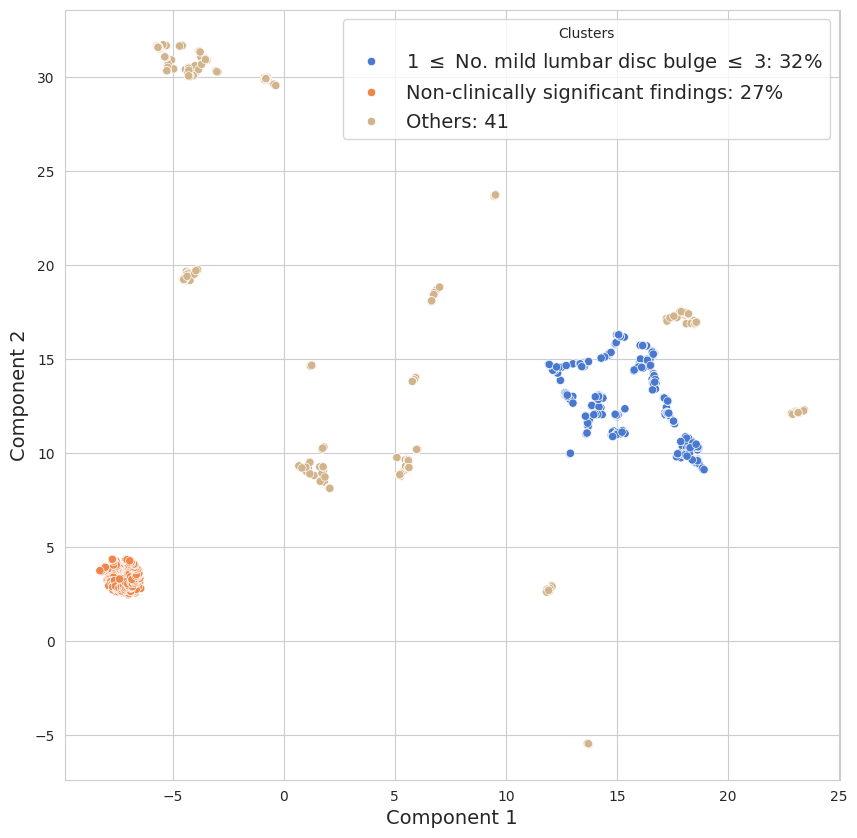}
        \caption{30-year-old age bracket}
    \end{subfigure}
    \begin{subfigure}[b]{0.47\textwidth}
        \centering
        \includegraphics[width=0.82\textwidth]{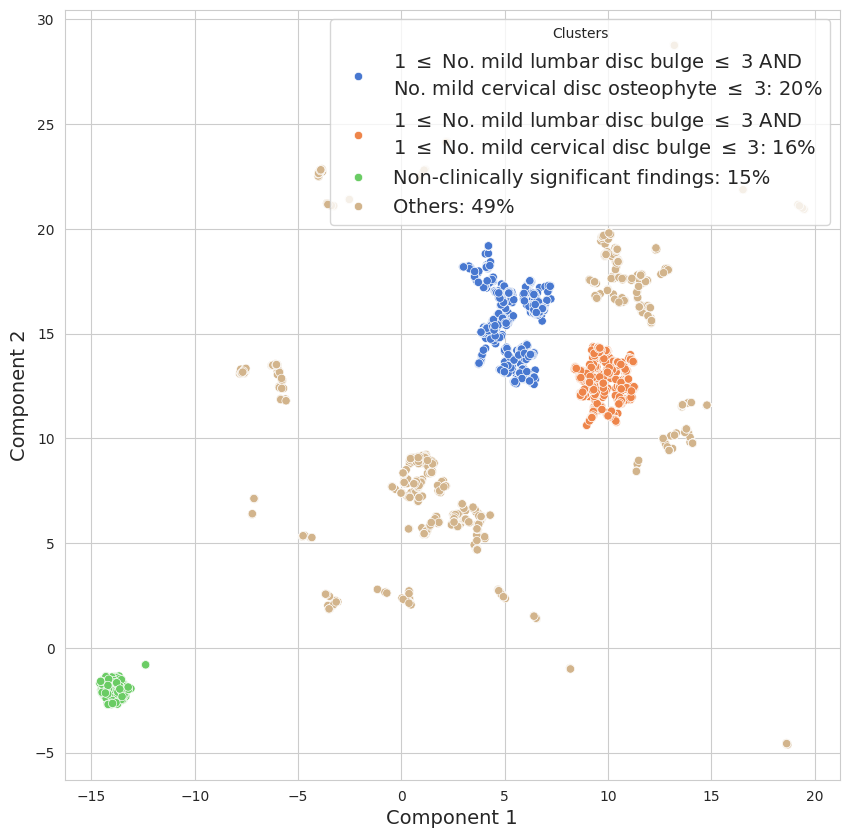}
        \caption{40-year-old age bracket}
    \end{subfigure}
    \begin{subfigure}[b]{0.47\textwidth}
        \centering
        \includegraphics[width=0.82\textwidth]{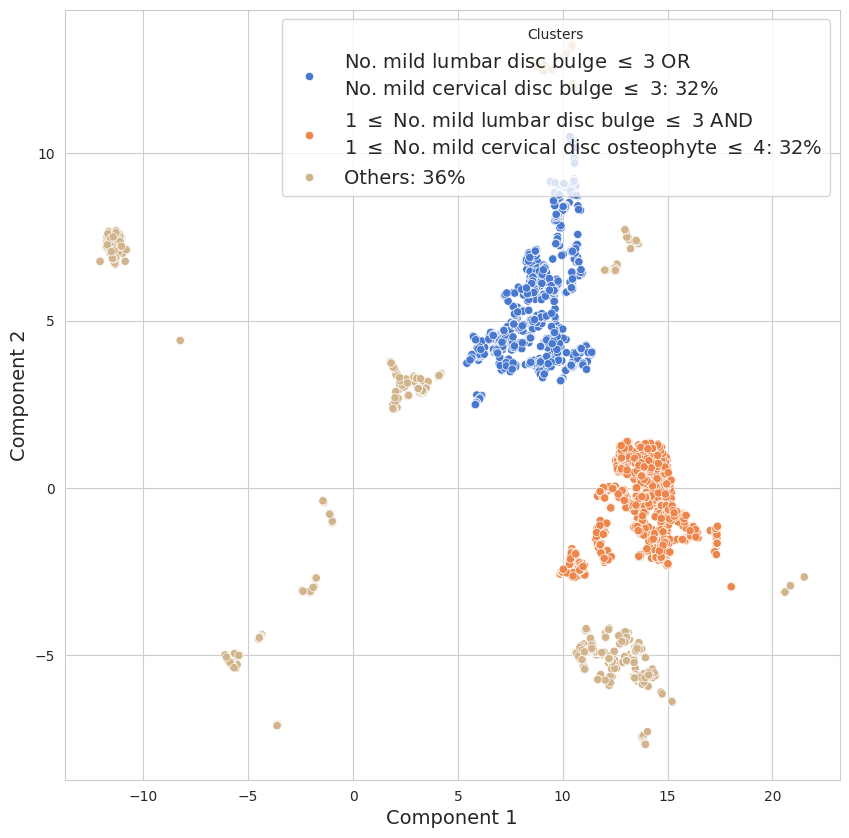}
        \caption{50-year-old age bracket}
    \end{subfigure}
    \begin{subfigure}[b]{0.47\textwidth}
        \centering
        \includegraphics[width=0.82\textwidth]{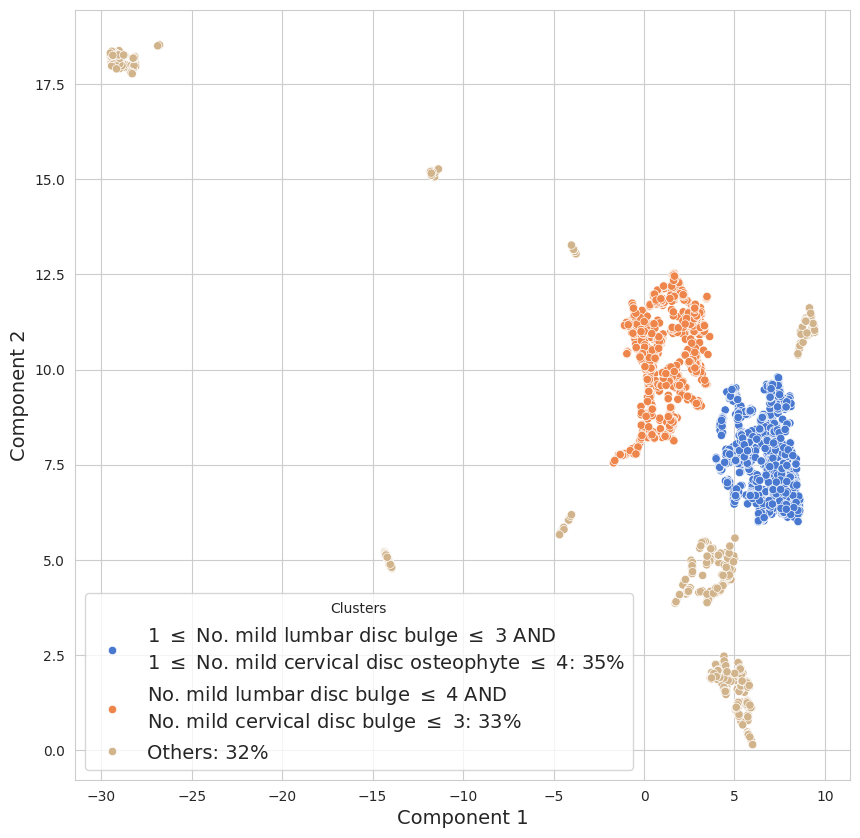}
        \caption{60-year-old age bracket}
    \end{subfigure}
    \begin{subfigure}[b]{0.47\textwidth}
        \centering
        \includegraphics[width=0.82\textwidth]{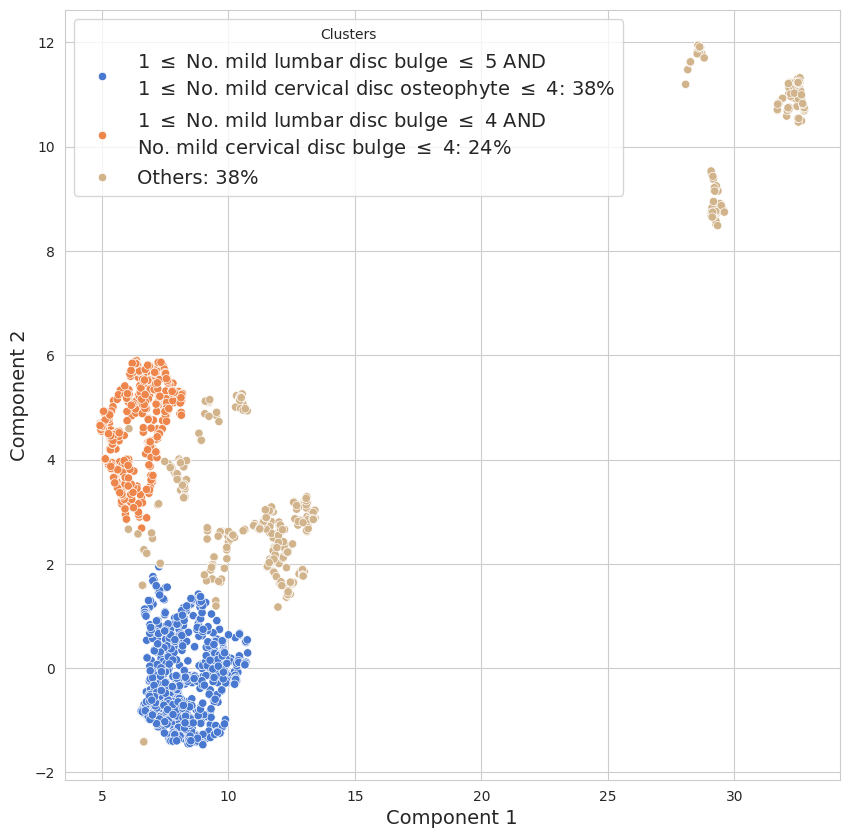}
        \caption{70-year-old age bracket}
    \end{subfigure}
    \begin{subfigure}[b]{0.47\textwidth}
        \centering
        \includegraphics[width=0.82\textwidth]{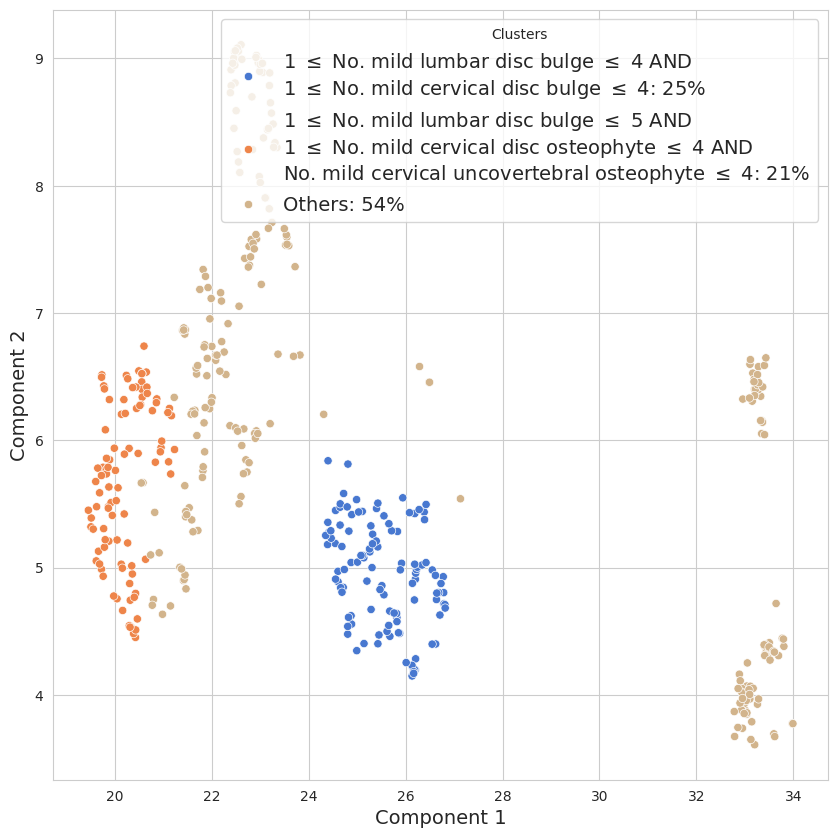}
        \caption{80-year-old age bracket}
    \end{subfigure}
    \caption{Clusters based on UMAP-reduced spine conditions. Only clusters that were more than 15\% of the population were kept and the rest merged into one. We labeled the clusters based on the dominant conditions in the cluster. The population percentage of each cluster is written in front of the label.}
    \label{fig_clusters}
\end{figure}

Our 15\% population threshold for age-based clusters described in Section \ref{sec:methods_normal} resulted in 32\% to 54\% of spines defined as abnormal across age brackets, as shown in Figure~\ref{fig_clusters}. We labeled normal clusters based on the primary conditions that were observed in the majority of samples in the cluster.

In the 30-year-old age bracket, having one to three mild lumbar disc bulges is the dominant cluster (32\% of the population). The next cluster represents 27\% of the population where most participants did not have a clinically significant finding. In the 40-year-old population, the population percentage of mild lumbar disc bulges is increased but divided into two groups based on having less or equal to 3 mild cervical disc osteophytes or between 1 to 3 mild cervical disc bulges. Moreover, the percentage of the population without a clinically significant finding reduced from 27\% to 15\%.

The two clusters based on mild lumbar disc bulges either with mild cervical disc osteophyte or mild cervical disc bulge exist in the 50-, 60-, and 70-year-old brackets as well. The only difference is the number of occurrences in  normal spines increases with age. Furthermore, the non-clinically significant findings cluster is not considered normal in these age groups. Finally, in the 80-year-old group, mild cervical uncovertebral osteophyte is another condition in addition to mild lumbar disc bulge and mild cervical disc osteophyte.

\subsection{Model Results}
\begin{table}
\caption{Comparison of age estimation models on normal data based on Mean Absolute Error (MAE) in years, $R^2$, and age bracket weighted MAE (WMAE) with and without Bias Correction (BC). Bold numbers represent the best results, while underlined numbers represent second-best results.}
\label{tab_ablation}
\centering
\begin{tabular}{cc|ccc|ccc}
\hline
\multicolumn{2}{c}{\multirow{2}{*}{Model}} & \multicolumn{3}{|c|}{Without BC} & \multicolumn{3}{|c}{With BC} \\
& & MAE & $R^2$ & WMAE & MAE & $R^2$ & WMAE \\
\hline
\multirow{2}{*}{Data size} & 85 samples & 9.77 & -0.03 & 11.51 & 9.76 & -0.04 & 10.55 \\
& 850 samples & 4.82 & 0.75 & 5.57 & 5.33 & 0.69 & 5.50 \\
\hline
Loss & Smooth-L1 & \underline{3.60} & \underline{0.86} & \underline{4.47} & \underline{3.94} & \underline{0.83} & \underline{4.00} \\
\hline
\multirow{3}{*}{Regions} & Cervical & 4.93 & 0.75 & 5.68 & 5.57 & 0.65 & 5.61 \\
& Thoracic & 4.02 & 0.82 & 4.77 & 4.57 & 0.77 & 4.58\\
& Lumbar & 3.89 & 0.83 & 4.67 & 4.35 & 0.79 & 4.57\\
\hline
\multicolumn{2}{c|}{Proposed model} & \textbf{3.47} & \textbf{0.87} & \textbf{3.60} & \textbf{3.67} & \textbf{0.85} & \textbf{3.60} \\
\hline
\end{tabular}
\end{table}

Model results comparing different age estimation models are shown in Table \ref{tab_ablation}. These results are based on model evaluation of the 1069 normal spine series in the test set. While looking at the performance of the models with and without bias correction in Table \ref{tab_ablation}, we note that although bias correction has increased the MAE, it has reduced the WMAE for all models except the proposed model where the WMAE remained unchanged. We see this because our data has an imbalanced age distribution with many cases close to the overall mean age in our sample (Figure \ref{fig_split}). Uncorrected predictions tend to be biased towards these mean ages, while bias correction removes the bias toward the mean and makes the error uniform. As a result bias correction increased the MAE, but the WMAE that computes error uniformly and independent of the imbalanced distribution across age brackets, declined. This result suggests that performing bias correction improved age estimation for our DCNN model. For the remainder of this manuscript, the spine age estimation refers to bias corrected values.

\subsubsection{Size of Dataset}
To evaluate the effect of dataset size on model performance, we tested our model on subsets of 85 and 850 samples in addition to the full dataset of 8491 series used for training the proposed model. The 85 samples subset represents 1\% of the data and is larger than the largest dataset previously used for predicting spine age~\cite{khan2013neural, sneath2022objective}. The 850-sample model represents the performance using 10\% of the data. It can be observed in Table \ref{tab_ablation} that increasing the training set size improves performance. Training with 85, 850, and 8491 series resulted in $R^2$ of $-0.04$, $0.69$, and $0.85$, respectively.

\subsubsection{Choice of the Loss Function}
We trained the proposed DCNN model using the MSE (L2) loss vs. smooth-L1 loss and compared their respective $R^2$ on the test set. The results of this experiment are shown in Table~\ref{tab_ablation}. L2 loss showed a $0.02$ improvement in $R^2$ and a reduction of $0.27$ and $0.40$ years in MAE and WMAE, respectively when compared to smooth-L1 loss.

\subsubsection{Region Specific vs. Whole Spine Model}
\textit{Are specific regions of the spine a better indicator of spine age?} In order to answer this question, we trained our model on masked inputs from different regions and compared their performance against the proposed whole spine model. The segmentation model~\cite{Khallaghi2023quantitative} is capable of detecting lumbar, thoracic, and cervical vertebrae. We used this feature to keep only one of these regions and mask the rest of the image. As seen in Table~\ref{tab_ablation}, the lumbar region exhibits improved performance compared to cervical and thoracic. This shows that aging is more apparent in the lumbar back region \cite{leboeuf2009pain}. Furthermore, the model trained on the whole spine region shows an improved performance of $0.06$ in $R^2$ compared to the model trained only on the lumbar region. This suggests that all regions of the spine are important for the assessment of the biological spine age.

\subsubsection{Repeat Scan Stability}

We evaluated model predictions on the 303 individuals who received two scans in the 8528 series in our test set. We expect repeat scans done on the same individual in a short time frame to have similar predicted ages. Results of these multiple scan predictions are shown in Table \ref{tab_scan_rescan}. We see moderately strong stability in our spine age predictions between scans, with an intraclass correlation of 0.73. We note that we do not expect perfect stability given there was an average of 1.6 years between scans and we expect biological and chronological aging differences between scans.

\begin{table}
\caption{Average years between scans, SAG intraclass correlation coefficient (ICC) and 95\% bootstrap confidence intervals of patients with two scans in our cohort.}
\label{tab_scan_rescan}
\centering
\begin{tabular}{c|ccc}
\hline
Subgroup & n & Avg Years & ICC \\
\hline
Full test set & 303 & 1.59 & 0.73 (0.68, 0.78) \\
\hline
Male & 172 & 1.60 & 0.72 (0.64, 0.79) \\
Female & 131 & 1.56 & 0.74 (0.63, 0.81) \\
\hline
\end{tabular}
\end{table}

\subsection{Analysis of Large Spine Age Differences}
\begin{figure}[!t]
\centerline{\includegraphics[width=\textwidth]{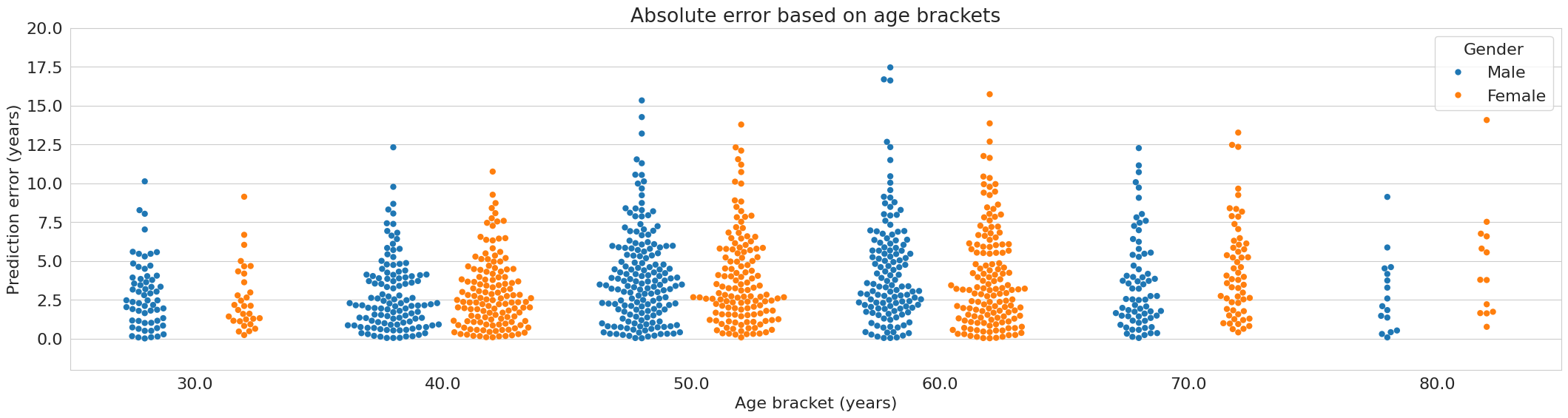}}
\caption{Absolute error on the normal test set grouped based on gender and age bracket}
\label{fig_swarm}
\end{figure}

In Figure~\ref{fig_swarm}, we illustrate the absolute error of the normal test set in years based on gender and age brackets. Upon conducting inference on the full-test set, we identified 30 instances where the discrepancy between chronological age and spine age exceeded 15 years. A thorough examination of these cases, supplemented by radiology reports and expert feedback, revealed that 23 of those cases were consistent with expected clinical patterns. For instance, we had a healthy 77-year-old participant whose spine age was 37. Of the remaining 7 samples, 2 of them had artifacts in the MR images, the segmentation model failed in 1 of the instances to properly mask the spine, and the spine age estimation model failed in 4 cases. The failure description is as follows:
\begin{enumerate}
    \item A 77-year-old participant whose spine age was estimated to be 39 years. However, based on expert feedback a spine age estimation between 45 to 50 years appears to be more consistent with the participant's spine condition. There are areas of increased intensity, such as C6 vertebra. Looking at the Grad-CAM the model has a high attention on C6 but was unable to predict the spine age more accurately.
    \item A 80-year-old participant with a vertebral fracture whose spine age was estimated at 63 years. The model has mostly focused on the disc bulges and has failed to attend to the vertebral fracture.
    \item A 57-year-old participant whose spine age was estimated at 74 years. There is a loss of thoracic and cervical curvature to some degree but there is nothing specific to explain the age overestimation.
    \item A 61-year-old participant whose spine age was estimated at 77 years. There is loss of intervertebral disc space at upper thoracic levels and slight rightward scoliosis but nothing special to indicate the overestimated spine age. After looking into Grad-CAM heatmaps, we realized that the model has focused on regions not specific to the spine, such as areas near the brain. 
\end{enumerate}

\subsection{Grad-CAM Results}\label{sec:gradcam}
\begin{figure}[!t]
\centerline{\includegraphics[width=\textwidth]{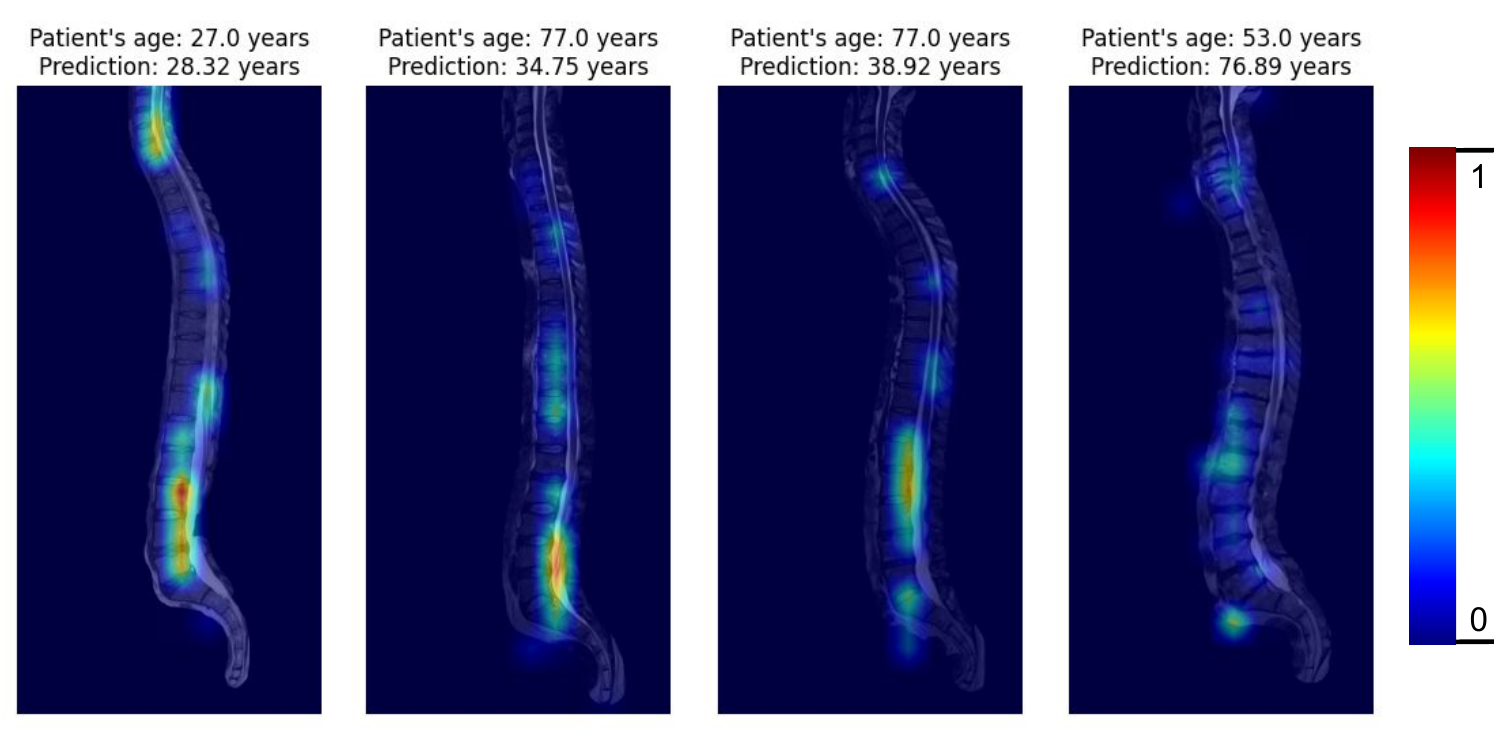}}
\caption{Grad-CAM heatmap on the middle frame of the MRI for four different patients. The values are adjusted based on $f(x) = max(ln(288x), 1)$ to have better contrast.}
\label{fig_gradcam}
\end{figure}

To interpret the model predictions, we used Grad-CAM \cite{selvaraju2017grad} to assign heatmaps to each region based on the contribution of the region in the final output. We generated two-dimensional heatmaps based on the fifth block of the model. The heatmaps are adjusted for a better contrast based on $f(x) = max(ln(288x), 1)$. The output of the function is visualized on the middle frame of the series in Figure~\ref{fig_gradcam}. It is important to note that the volume has been reduced to one channel at the fifth block, so it is not possible to generate a 3D Grad-CAM heatmap at this layer.

Across all images in Figure~\ref{fig_gradcam}, it can be seen that the model focuses on the disc bulges as one of the main indicators of aging. The second image from the left shows a relatively healthy spine for a 77 year old, with a predicted spine age of 34.75 years. The third image is the series the model has attended to C6 as radiologists have suggested, but focused more on disc bulges. Based on radiology feedback, the spine age should be younger than the chronological age but around 45 to 50 years, not 39 years. The last picture depicts a participant whose spine has degenerated significantly compared to their chronological age.

\subsection{Spine Age Gap as a Biomarker for Spine Health}\label{sec:biomarker}

In this section, we explore the relationship between spine conditions, lifestyle factors, and predicted spine age. Specifically, we aim to determine whether the gap between chronological age and predicted spine age is associated with clinically relevant spine conditions, and whether this spine age gap can be used as a biomarker for overall spine health. 

We begin by quantifying the associations between SAG and spinal conditions in Table \ref{tab_regression}. We fit separate linear regression models using lumbar degenerative conditions, spinal structural and canal pathologies, or lifestyle factors as covariates, with SAG as the predicted outcome. Lumbar degenerative conditions were categorized and in conditions with multiple severities cases were only assigned to the most severe category present. We observe that lumbar disc bulges and osteophytes are significantly associated with SAG, with the occurrence of a severe disc bulge resulting in a predicted increased SAG of $2.96$ years on average. Several spinal and canal pathologies are also associated with an increased SAG, including spinal stenosis, fractures, and spinal canal narrowing. We find similar associations between SAG and cervical degenerative conditions (see the Appendix for details).

Examining the association between SAG and lifestyle factors, we find higher levels of smoking and physically demanding work are associated with an increase in SAG, while vigorous exercise is associated with a decrease in SAG. This association with between spine health and smoking aligns with previous studies \cite{akmal2004effect,berman2017effect} that identified an association between smoking and intervertebral disc degeneration, as well as changes in vertebral bone structure. Alcohol consumption has only a mild association with SAG, which again aligns with previous research indicating no substantial relationship between alcohol intake and spinal degeneration \cite{gorman1987relationship, leboeuf2000alcohol, holmberg2006primary, lv2018prevalence}.

\begin{table*}
\caption{Linear regression coefficients and 95\% confidence intervals (CIs) quantifying the association between lumbar degenerative conditions, spine structural and canal pathologies, and lifestyle factors with spine-age gap. Separate regression models were fit for lumbar degenerate conditions, spinal structural pathologies, and lifestyle factors. In all cases models controlled for biological sex.}
\label{tab_regression}
\centering
\begin{tabular}{ll|c|c}
\hline
\multicolumn{2}{c|}{Condition} & n & Effect (95\% CI) \\
\hline
\multirow{14}{*}{\shortstack{Lumbar degenerative \\ spinal conditions}} 
& Mild disc bulge (No.$>$2) & 801 & 1.27 (0.92, 1.63)$^*$\\
& Moderate disc bulge (No.$>$1) & 85 & 1.58 (0.55, 2.61)$^*$ \\
& Severe disc bulge (No.$>$0) & 12 & 2.96 (0.27, 5.65)$^*$ \\
& Mild disc dessication (No.$>$1) & 315 & -0.30 (-0.84, 0.25) \\
& Moderate disc dessication (No.$>$0) & 295 & 0.02 (-0.55, 0.59) \\
& Severe disc dessication (No.$>$0) & 212 & 0.20 (-0.48, 0.88) \\
& Near complete disc dessication (No.$>$0) & 66 & -0.59 (-1.76, 0.59) \\
& Disc annular fissure (No.$>$0) & 781 & 0.24 (-0.12, 0.59) \\
& Vertebral endplate change (No.$>$0) & 547 & 0.19 (-0.26, 0.64) \\
& Mild disc osteophyte (No.$>$1) & 630 & 2.34 (1.95, 2.73)$^*$ \\
& Moderate disc osteophyte (No.$>$0) & 36 & 2.46 (0.88, 4.03)$^*$ \\
& Mild disc protrusion (No.$>$1) & 249 & 0.41 (-0.19, 1.02) \\
& Moderate disc protrusion (No.$>$0) & 94 & 0.54 (-0.43, 1.51) \\
& Mild disc extrusion (No.$>$0) & 71 & -0.59 (-1.70, 0.52) \\
\hline
\multirow{10}{*}{\shortstack{Spinal structural and\\canal pathologies}} 
& Spondylolisthesis & 722 & 0.91 (0.56, 1.26)$^*$ \\
& Scoliosis & 1477 & 0.64 (0.37, 0.91)$^*$ \\
& Kyphosis or lordosis & 3061 & 0.45 (0.23, 0.66)$^*$ \\
& Fracture & 229 & 1.45 (0.82, 2.07)$^*$ \\
& Spinal stenosis & 125 & 1.87 (1.03, 2.72)$^*$ \\
& Congenital spinal canal narrowing & 197 & 1.18 (0.50, 1.85)$^*$ \\
& Cord abnormalities & 194 & 0.22 (-0.46, 0.90) \\
& Transitional vertebra & 542 & -0.07 (-0.49, 0.34) \\
& Tralov perineural cyst & 601 & -0.30 (-0.70, 0.10) \\
\hline
\multirow{7}{*}{Lifestyle factors}
& Packs per day smoked $^\ddagger$ & 2302 & 0.93 (0.64, 1.22)$^*$  \\
& Days per week consuming alcohol $^\ddagger$ & 6136 & 0.08 (0.03, 0.13)$^*$  \\
& Time sedentary $^\ddagger$ & 8332 & -0.01 (-0.04, 0.03) \\
& Physically moderate work & 1001 & 0.29 (-0.03, 0.62)  \\
& Physically heavy work & 488 &  0.67 (0.22, 1.12)$^*$ \\
& Moderate exercise & 2146 & -0.40 (-0.70, -0.11)$^*$ \\
& Vigorous exercise & 4285 &  -0.79 (-1.05, -0.52)$^*$ \\
\hline
\multicolumn{4}{l}{\footnotesize{$^*$ statistically significant effects. $^\ddagger$ continuous variables, with counts representing the number greater than zero.}}
    \end{tabular}
\end{table*}

\begin{figure}[!t]
\centerline{\includegraphics[width=0.5\textwidth]{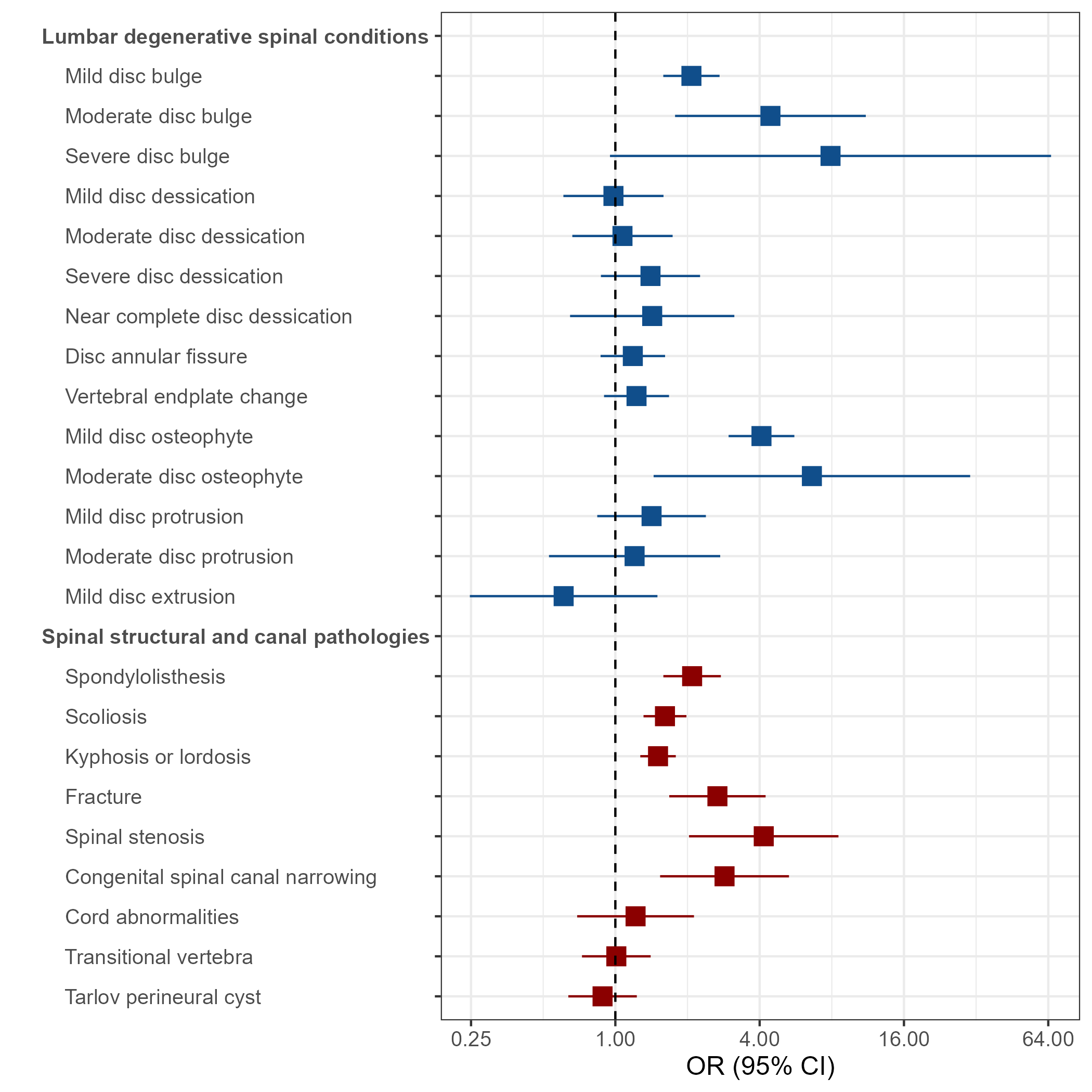}}
\caption{Odds ratios (OR) for lumbar degenerative and spinal structural conditions comparing cases with large positive spine age gaps (greater than 5 years) and large negative spine age gaps (less than -5 years). The x-axis depicts odds ratios between large and small spine age groups, with horizontal lines representing 95\% confidence intervals for the ORs.}
\label{fig_large_SAG}
\end{figure}

Next, we explore whether large spine age gaps are associated with an increased likelihood of clinically relevant spinal conditions. We compare the odds of degenerative lumbar and spinal structural conditions in the test set of cases where individuals had a SAG of greater than 5, with cases where individuals had a SAG of less than -5 (Figure \ref{fig_large_SAG}). We find the odds of individuals having lumbar disc bulges or osteophytes much higher in the group with $\text{SAG}>5$. Individuals with a large positive SAG have odds of moderate disc bulges four-times higher than individuals with a large negative SAG, and this grows to eight-fold higher for severe disc bulges. Similarly, the odds of spinal fractures, stenosis, spondylolisthesis, and canal narrowing are two- to four-times larger in individuals with a large positive SAG. 

\begin{figure}[!t]
\centerline{\includegraphics[width=0.5\textwidth]{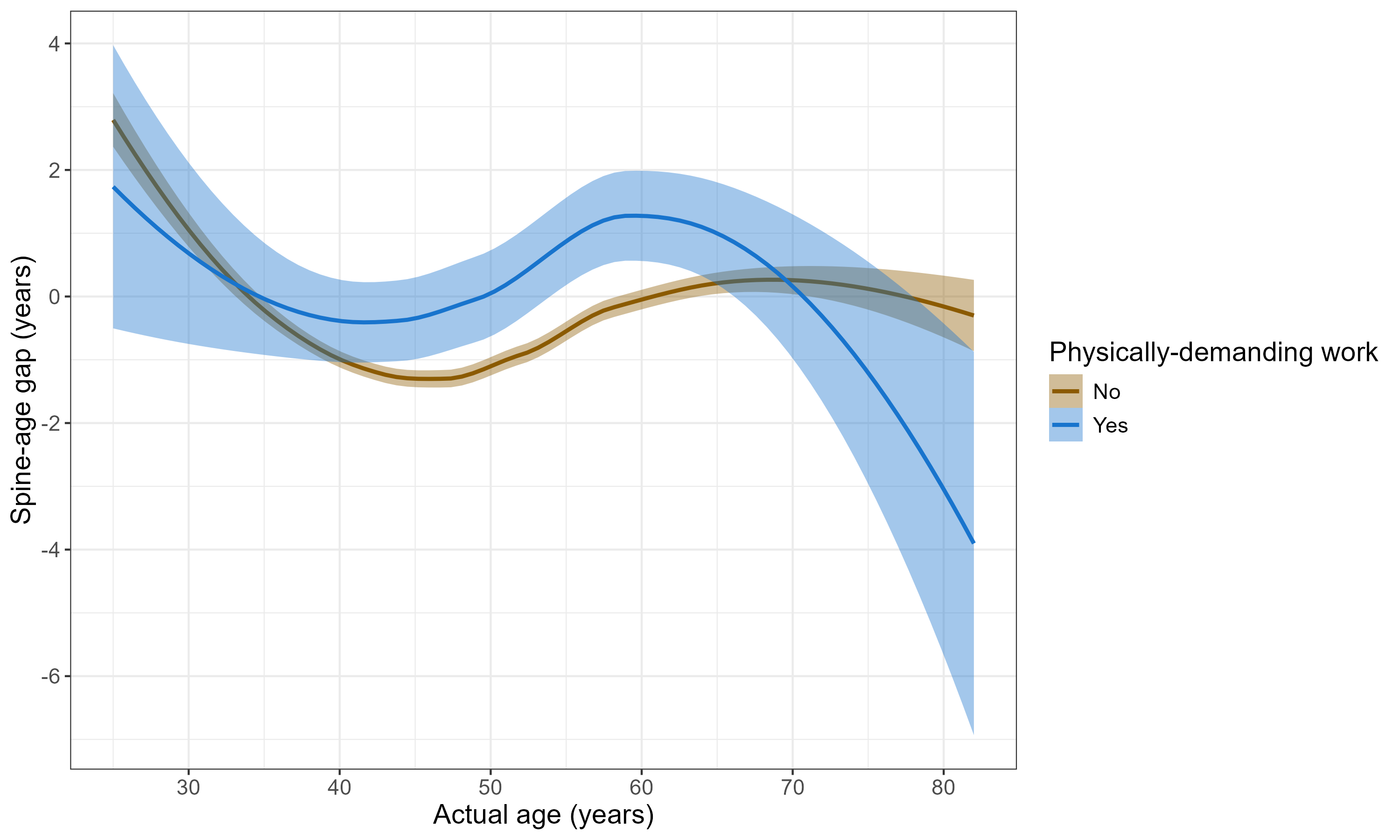}}
\caption{Relationship between spine-age gap and actual age in individuals who do and do not perform physically demanding work.}
\label{fig_work_heavy}
\end{figure}

The relationship between spine health and lifestyle factors may change over chronological age. For example, it may be that performing physically demanding work degrades spine health when individuals are young. However, for older individuals, the ability to perform physically demanding work at all may suggest better spine health relative to those not performing such work. This reverse in relationship over time is exactly what we see when comparing SAG and actual age in individuals who do and do not perform physically demanding work (Figure \ref{fig_work_heavy}). Average SAG is higher for younger individuals performing physically demanding work, but lower for older individuals. This shows that SAG may be a relevant marker of overall spine health.  

\section{Conclusion}

We believe this work is a significant step towards understanding spine health and its key factors. By analyzing radiology reports, we developed a data-driven framework for detecting typical age-related spine degradation in different age groups, assessed alternate approaches to achieve optimal model performance for predicting spine age, and demonstrated that SAG is an indicator of spine health based on its associations with spinal conditions and lifestyle factors.

This study is the first to utilize deep learning and convolutional neural networks to estimate spine age based on sagittal T2-weighted MRI series. We developed a network inspired by brain age estimation research to estimate spine age. Our experiments were conducted on a vast dataset of over 18,000 MRI series and 17,000 participants, with a detailed study on the clinical relevance of SAG and spine conditions in more than 8,500 MRI series.

One limitation of our work is the lack of data on rare or severe conditions as the scans were primarily for preventive health screening. Hospitals that deal with patients with these rare conditions may enhance model performance with additional data. For example, conditions like soft tissue edema and severe lumbar disc osteophyte were rare in our dataset, limiting our ability to analyze model performance on these conditions. More data may also help improve our robustness to spine fractures, which was missed in one of the failure cases.

Future work includes:
\begin{enumerate}
    \item Investigating the relationship between performance improvement and clinical findings by using more data, especially data with rare or severe spine conditions.
    \item In our current model, we utilized a Deep Convolutional Neural Network. There is potential to investigate state-of-the-art deep learning models, such as vision transformers, which could enhance both the model performance and the clinical findings.
    \item Additionally, we employed a combination of UMAP and HDBSCAN to identify normal and abnormal spine groups within age brackets. Future research could explore more effective methods to replace these techniques. For example, an encoder-decoder-based approach could be used for dimensionality reduction instead of UMAP, providing more meaningful features and potentially improving the identification of normal populations.
    \item Furthermore, our technique of identifying normal participants, estimating age, and analyzing clinical relevance can be applied to other organs, such as prostate, kidney, and liver.
\end{enumerate}

\bibliographystyle{unsrt}  
\bibliography{references}  

@inproceedings{khan2013neural,
  title={Neural Network Based Spinal Age Estimation Using Lumbar Spine Magnetic Resonance Images {(MRI)}},
  author={Khan, Atif and Iliescu, Daciana and Hines, Evor and Hutchinson, Charles and Sneath, Robert},
  booktitle={2013 4th International Conference on Intelligent Systems, Modelling and Simulation},
  pages={88--93},
  year={2013},
  organization={IEEE}
}

@article{watanabe2006age,
  title={Age estimation from the degree of osteophyte formation of vertebral columns in Japanese},
  author={Watanabe, Satoshi and Terazawa, Koichi},
  journal={Legal medicine},
  volume={8},
  number={3},
  pages={156--160},
  year={2006},
  publisher={Elsevier}
}

@article{ruhli2005age,
  title={Age-dependent changes of the normal human spine during adulthood},
  author={R{\"u}hli, Frank J and M{\"u}ntener, M and Henneberg, M},
  journal={American Journal of Human Biology: The Official Journal of the Human Biology Association},
  volume={17},
  number={4},
  pages={460--469},
  year={2005},
  publisher={Wiley Online Library}
}

@article{sneath2022objective,
  title={An Objective Assessment of Lumbar Spine Degeneration/Ageing Seen on {MRI} Using An Ensemble Method—A Novel Approach to Lumbar {MRI} Reporting},
  author={Sneath, Robert JS and Khan, Atif and Hutchinson, Charles},
  journal={Spine},
  volume={47},
  number={5},
  pages={E187--E195},
  year={2022},
  publisher={LWW}
}

@article{mcinnes2018umap,
  title={Umap: Uniform manifold approximation and projection for dimension reduction},
  author={McInnes, Leland and Healy, John and Melville, James},
  journal={arXiv preprint arXiv:1802.03426},
  year={2018}
}

@article{mcinnes2017hdbscan,
  title={hdbscan: Hierarchical density based clustering.},
  author={McInnes, Leland and Healy, John and Astels, Steve and others},
  journal={J. Open Source Softw.},
  volume={2},
  number={11},
  pages={205},
  year={2017}
}

@misc{leonardsen2022deep,
  title={Deep neural networks learn general and clinically relevant representations of the ageing brain. NeuroImage, 256, Article 119210},
  author={Leonardsen, EH and Peng, H and Kaufmann, T and Agartz, I and Andreassen, OA and Celius, EG and Espeseth, T and Harbo, HF and H{\o}gest{\o}l, EA and de Lange, AM and others},
  year={2022}
}

@inproceedings{lee2024smoking,
  title={Smoking Is Associated with Accelerated Brain Aging: A Large, Diverse Population-based Analysis (S43. 003)},
  author={Lee, Soojin and Garg, Saurabh and Datta, Madhurima and Nguyen, Duc and Akbari, Nasrin and Rajendran, Arun and Hashemi, Sam and Chodakiewitz, Yosef and Attariwala, Rajpaul},
  booktitle={Neurology},
  volume={102},
  number={17\_supplement\_1},
  pages={2376},
  year={2024},
  organization={AAN Enterprises}
}

@inproceedings{selvaraju2017grad,
  title={Grad-cam: Visual explanations from deep networks via gradient-based localization},
  author={Selvaraju, Ramprasaath R and Cogswell, Michael and Das, Abhishek and Vedantam, Ramakrishna and Parikh, Devi and Batra, Dhruv},
  booktitle={Proceedings of the IEEE international conference on computer vision},
  pages={618--626},
  year={2017}
}

@inproceedings{shen2018deep,
  title={Deep regression forests for age estimation},
  author={Shen, Wei and Guo, Yilu and Wang, Yan and Zhao, Kai and Wang, Bo and Yuille, Alan L},
  booktitle={Proceedings of the IEEE conference on computer vision and pattern recognition},
  pages={2304--2313},
  year={2018}
}

@inproceedings{zhang2017age,
  title={Age progression/regression by conditional adversarial autoencoder},
  author={Zhang, Zhifei and Song, Yang and Qi, Hairong},
  booktitle={Proceedings of the IEEE conference on computer vision and pattern recognition},
  pages={5810--5818},
  year={2017}
}

@inproceedings{shu2015personalized,
  title={Personalized age progression with aging dictionary},
  author={Shu, Xiangbo and Tang, Jinhui and Lai, Hanjiang and Liu, Luoqi and Yan, Shuicheng},
  booktitle={Proceedings of the IEEE international conference on computer vision},
  pages={3970--3978},
  year={2015}
}

@inproceedings{kemelmacher2014illumination,
  title={Illumination-aware age progression},
  author={Kemelmacher-Shlizerman, Ira and Suwajanakorn, Supasorn and Seitz, Steven M},
  booktitle={Proceedings of the IEEE conference on computer vision and pattern recognition},
  pages={3334--3341},
  year={2014}
}

@article{savage1997relationship,
  title={The relationship between the magnetic resonance imaging appearance of the lumbar spine and low back pain, age and occupation in males},
  author={Savage, RA and Whitehouse, GH and Roberts, N},
  journal={European Spine Journal},
  volume={6},
  pages={106--114},
  year={1997},
  publisher={Springer}
}

@article{zhang2023age,
  title={Age-level bias correction in brain age prediction},
  author={Zhang, Biao and Zhang, Shuqin and Feng, Jianfeng and Zhang, Shihua},
  journal={NeuroImage: Clinical},
  volume={37},
  pages={103319},
  year={2023},
  publisher={Elsevier}
}

@article{cole2018brain,
  title={Brain age predicts mortality},
  author={Cole, James H and Ritchie, Stuart J and Bastin, Mark E and Hern{\'a}ndez, Vald{\'e}s and Mu{\~n}oz Maniega, S and Royle, Natalie and Corley, Janie and Pattie, Alison and Harris, Sarah E and Zhang, Qian and others},
  journal={Molecular psychiatry},
  volume={23},
  number={5},
  pages={1385--1392},
  year={2018},
  publisher={Nature Publishing Group}
}

@inproceedings{shah2024ordinal,
  title={Ordinal Classification with Distance Regularization for Robust Brain Age Prediction},
  author={Shah, Jay and Siddiquee, Md Mahfuzur Rahman and Su, Yi and Wu, Teresa and Li, Baoxin},
  booktitle={Proceedings of the IEEE/CVF Winter Conference on Applications of Computer Vision},
  pages={7882--7891},
  year={2024}
}

@article{shen2022attention,
  title={Attention-guided deep learning for gestational age prediction using fetal brain {MRI}},
  author={Shen, Liyue and Zheng, Jimmy and Lee, Edward H and Shpanskaya, Katie and McKenna, Emily S and Atluri, Mahesh G and Plasto, Dinko and Mitchell, Courtney and Lai, Lillian M and Guimaraes, Carolina V and others},
  journal={Scientific reports},
  volume={12},
  number={1},
  pages={1408},
  year={2022},
  publisher={Nature Publishing Group UK London}
}

@article{jonsson2019brain,
  title={Brain age prediction using deep learning uncovers associated sequence variants},
  author={J{\'o}nsson, Benedikt Atli and Bjornsdottir, Gyda and Thorgeirsson, TE and Ellingsen, Lotta Mar{\'\i}a and Walters, G Bragi and Gudbjartsson, DF and Stefansson, Hreinn and Stefansson, Kari and Ulfarsson, MO},
  journal={Nature communications},
  volume={10},
  number={1},
  pages={5409},
  year={2019},
  publisher={Nature Publishing Group UK London}
}

@article{lee2022deep,
  title={Deep learning-based brain age prediction in normal aging and dementia},
  author={Lee, Jeyeon and Burkett, Brian J and Min, Hoon-Ki and Senjem, Matthew L and Lundt, Emily S and Botha, Hugo and Graff-Radford, Jonathan and Barnard, Leland R and Gunter, Jeffrey L and Schwarz, Christopher G and others},
  journal={Nature Aging},
  volume={2},
  number={5},
  pages={412--424},
  year={2022},
  publisher={Nature Publishing Group US New York}
}

@article{de2022mind,
  title={Mind the gap: Performance metric evaluation in brain-age prediction},
  author={de Lange, Ann-Marie G and Anat{\"u}rk, Melis and Rokicki, Jaroslav and Han, Laura KM and Franke, Katja and Aln{\ae}s, Dag and Ebmeier, Klaus P and Draganski, Bogdan and Kaufmann, Tobias and Westlye, Lars T and others},
  journal={Human Brain Mapping},
  volume={43},
  number={10},
  pages={3113--3129},
  year={2022},
  publisher={Wiley Online Library}
}

@article{peng2021accurate,
  title={Accurate brain age prediction with lightweight deep neural networks},
  author={Peng, Han and Gong, Weikang and Beckmann, Christian F and Vedaldi, Andrea and Smith, Stephen M},
  journal={Medical image analysis},
  volume={68},
  pages={101871},
  year={2021},
  publisher={Elsevier}
}

@article{armanious2021age,
  title={Age-net: An {MRI}-based iterative framework for brain biological age estimation},
  author={Armanious, Karim and Abdulatif, Sherif and Shi, Wenbin and Salian, Shashank and K{\"u}stner, Thomas and Weiskopf, Daniel and Hepp, Tobias and Gatidis, Sergios and Yang, Bin},
  journal={IEEE Transactions on Medical Imaging},
  volume={40},
  number={7},
  pages={1778--1791},
  year={2021},
  publisher={IEEE}
}

@article{cheng2021brain,
  title={{Brain age estimation from MRI using cascade networks with ranking loss}},
  author={Cheng, Jian and Liu, Ziyang and Guan, Hao and Wu, Zhenzhou and Zhu, Haogang and Jiang, Jiyang and Wen, Wei and Tao, Dacheng and Liu, Tao},
  journal={IEEE Transactions on Medical Imaging},
  volume={40},
  number={12},
  pages={3400--3412},
  year={2021},
  publisher={IEEE}
}

@article{zhang2022robust,
  title={Robust brain age estimation based on {sMRI} via nonlinear age-adaptive ensemble learning},
  author={Zhang, Zhaonian and Jiang, Richard and Zhang, Ce and Williams, Bryan and Jiang, Ziping and Li, Chang-Tsun and Chazot, Paul and Pavese, Nicola and Bouridane, Ahmed and Beghdadi, Azeddine},
  journal={IEEE Transactions on Neural Systems and Rehabilitation Engineering},
  volume={30},
  pages={2146--2156},
  year={2022},
  publisher={IEEE}
}

@article{cole2017predicting,
  title={Predicting brain age with deep learning from raw imaging data results in a reliable and heritable biomarker},
  author={Cole, James H and Poudel, Rudra PK and Tsagkrasoulis, Dimosthenis and Caan, Matthan WA and Steves, Claire and Spector, Tim D and Montana, Giovanni},
  journal={NeuroImage},
  volume={163},
  pages={115--124},
  year={2017},
  publisher={Elsevier}
}

@article{simonyan2014very,
  title={Very deep convolutional networks for large-scale image recognition},
  author={Simonyan, Karen and Zisserman, Andrew},
  journal={arXiv preprint arXiv:1409.1556},
  year={2014}
}

@inproceedings{szegedy2015going,
  title={Going deeper with convolutions},
  author={Szegedy, Christian and Liu, Wei and Jia, Yangqing and Sermanet, Pierre and Reed, Scott and Anguelov, Dragomir and Erhan, Dumitru and Vanhoucke, Vincent and Rabinovich, Andrew},
  booktitle={Proceedings of the IEEE conference on computer vision and pattern recognition},
  pages={1--9},
  year={2015}
}

@inproceedings{he2016deep,
  title={Deep residual learning for image recognition},
  author={He, Kaiming and Zhang, Xiangyu and Ren, Shaoqing and Sun, Jian},
  booktitle={Proceedings of the IEEE conference on computer vision and pattern recognition},
  pages={770--778},
  year={2016}
}

@article{Khallaghi2023quantitative,
  title={Quantitative Assessment of the Whole Spine in {T2 MRI} Using Deep Learning},
  author={Khallaghi, Siavash and Porto, Lucas and London, Sean and Chodakiewitz, Yosef and Attariwal, Rajpaul and Hashemi, Sam},
  journal={International Society for Magnetic Resonance in Medicine (ISMRM)},
  year={2023}
}

@article{papadakis2011pathophysiology,
  title={Pathophysiology and biomechanics of the aging spine},
  author={Papadakis, Michael and Sapkas, Georgios and Papadopoulos, Elias C and Katonis, Pavlos},
  journal={The open orthopaedics journal},
  volume={5},
  pages={335},
  year={2011},
  publisher={Bentham Science Publishers}
}

@article{leboeuf2009pain,
  title={Pain in the lumbar, thoracic or cervical regions: do age and gender matter? A population-based study of 34,902 {Danish} twins 20--71 years of age},
  author={Leboeuf-Yde, Charlotte and Nielsen, Jan and Kyvik, Kirsten O and Fejer, Ren{\'e} and Hartvigsen, Jan},
  journal={BMC musculoskeletal disorders},
  volume={10},
  pages={1--12},
  year={2009},
  publisher={Springer}
}

@article{gille2017new,
  title={A new classification system for degenerative spondylolisthesis of the lumbar spine},
  author={Gille, Olivier and Bouloussa, Houssam and Mazas, Simon and Vergari, Claudio and Challier, Vincent and Vital, Jean-Marc and Coudert, Pierre and Ghailane, Soufiane},
  journal={European Spine Journal},
  volume={26},
  pages={3096--3105},
  year={2017},
  publisher={Springer}
}

@article{pfirrmann2001magnetic,
  title={Magnetic resonance classification of lumbar intervertebral disc degeneration},
  author={Pfirrmann, Christian WA and Metzdorf, Alexander and Zanetti, Marco and Hodler, Juerg and Boos, Norbert},
  journal={spine},
  volume={26},
  number={17},
  pages={1873--1878},
  year={2001},
  publisher={LWW}
}

@article{gille2014degenerative,
  title={Degenerative lumbar spondylolisthesis. Cohort of 670 patients, and proposal of a new classification},
  author={Gille, O and Challier, V and Parent, H and Cavagna, R and Poignard, A and Faline, A and Fuentes, S and Ricart, O and Ferrero, E and Slimane, M Ould and others},
  journal={Orthopaedics \& Traumatology: Surgery \& Research},
  volume={100},
  number={6},
  pages={S311--S315},
  year={2014},
  publisher={Elsevier}
}

@article{riesenburger2015novel,
  title={A novel classification system of lumbar disc degeneration},
  author={Riesenburger, Ron I and Safain, Mina G and Ogbuji, Richard and Hayes, Jackson and Hwang, Steven W},
  journal={Journal of Clinical Neuroscience},
  volume={22},
  number={2},
  pages={346--351},
  year={2015},
  publisher={Elsevier}
}

@misc{kim1992mri,
  title={MRI classification of lumbar herniated intervertebral disc},
  author={Kim, Key Yong and Kim, Yung Tae and Lee, Choon Sung and Shin, Myung Jin},
  journal={Orthopedics},
  volume={15},
  number={4},
  pages={493--497},
  year={1992},
  publisher={SLACK Incorporated Thorofare, NJ}
}

@inproceedings{he2021spineone,
  title={SpineOne: A One-Stage Detection Framework for Degenerative Discs and Vertebrae},
  author={He, Jiabo and Liu, Wei and Wang, Yu and Ma, Xingjun and Hua, Xian-Sheng},
  booktitle={2021 IEEE International Conference on Bioinformatics and Biomedicine (BIBM)},
  pages={1331--1334},
  year={2021},
  organization={IEEE}
}

@article{chen2024deep,
  title={Deep Learning-Based Intelligent Diagnosis of Lumbar Diseases with Multi-Angle View of Intervertebral Disc},
  author={Chen, Kaisi and Zheng, Lei and Zhao, Honghao and Wang, Zihang},
  journal={Mathematics},
  volume={12},
  number={13},
  pages={2062},
  year={2024},
  publisher={MDPI}
}

@article{zheng2022deep,
  title={Deep learning-based high-accuracy quantitation for lumbar intervertebral disc degeneration from MRI},
  author={Zheng, Hua-Dong and Sun, Yue-Li and Kong, De-Wei and Yin, Meng-Chen and Chen, Jiang and Lin, Yong-Peng and Ma, Xue-Feng and Wang, Hong-Shen and Yuan, Guang-Jie and Yao, Min and others},
  journal={Nature communications},
  volume={13},
  number={1},
  pages={841},
  year={2022},
  publisher={Nature Publishing Group UK London}
}

@article{yi2023deep,
  title={Deep learning-based high-accuracy detection for lumbar and cervical degenerative disease on T2-weighted MR images},
  author={Yi, Wei and Zhao, Jingwei and Tang, Wen and Yin, Hongkun and Yu, Lifeng and Wang, Yaohui and Tian, Wei},
  journal={European Spine Journal},
  volume={32},
  number={11},
  pages={3807--3814},
  year={2023},
  publisher={Springer}
}

@inproceedings{lu2018deep,
  title={Deep spine: automated lumbar vertebral segmentation, disc-level designation, and spinal stenosis grading using deep learning},
  author={Lu, Jen-Tang and Pedemonte, Stefano and Bizzo, Bernardo and Doyle, Sean and Andriole, Katherine P and Michalski, Mark H and Gonzalez, R Gilberto and Pomerantz, Stuart R},
  booktitle={Machine Learning for Healthcare Conference},
  pages={403--419},
  year={2018},
  organization={PMLR}
}

@article{hallinan2021deep,
  title={Deep learning model for automated detection and classification of central canal, lateral recess, and neural foraminal stenosis at lumbar spine MRI},
  author={Hallinan, James Thomas Patrick Decourcy and Zhu, Lei and Yang, Kaiyuan and Makmur, Andrew and Algazwi, Diyaa Abdul Rauf and Thian, Yee Liang and Lau, Samuel and Choo, Yun Song and Eide, Sterling Ellis and Yap, Qai Ven and others},
  journal={Radiology},
  volume={300},
  number={1},
  pages={130--138},
  year={2021},
  publisher={Radiological Society of North America}
}

@article{akmal2004effect,
  title={Effect of nicotine on spinal disc cells: a cellular mechanism for disc degeneration},
  author={Akmal, Mohammed and Kesani, Anil and Anand, Bobby and Singh, Abhinav and Wiseman, Mike and Goodship, Allen},
  journal={Spine},
  volume={29},
  number={5},
  pages={568--575},
  year={2004},
  publisher={LWW}
}

@article{berman2017effect,
  title={The effect of smoking on spinal fusion},
  author={Berman, Daniel and Oren, Jonathan H and Bendo, John and Spivak, Jeffrey},
  journal={International journal of spine surgery},
  volume={11},
  number={4},
  year={2017},
  publisher={International Journal of Spine Surgery}
}

@article{gorman1987relationship,
  title={Relationship between alcohol abuse and low back pain},
  author={Gorman, DM and Potamianos, G and Williams, KA and Frank, AO and Duffy, SW and Peters, TJ},
  journal={Alcohol and Alcoholism},
  volume={22},
  number={1},
  pages={61--63},
  year={1987},
  publisher={Oxford University Press}
}

@article{leboeuf2000alcohol,
  title={Alcohol and low-back pain: a systematic literature review},
  author={Leboeuf-Yde, Charlotte},
  journal={Journal of Manipulative and Physiological Therapeutics},
  volume={23},
  number={5},
  pages={343--346},
  year={2000},
  publisher={Elsevier}
}

@article{holmberg2006primary,
  title={Primary care consultation, hospital admission, sick leave and disability pension owing to neck and low back pain: a 12-year prospective cohort study in a rural population},
  author={Holmberg, Sara AC and Thelin, Anders G},
  journal={BMC Musculoskeletal Disorders},
  volume={7},
  pages={1--8},
  year={2006},
  publisher={Springer}
}

@article{lv2018prevalence,
  title={The prevalence and associated factors of symptomatic cervical Spondylosis in Chinese adults: a community-based cross-sectional study},
  author={Lv, Yanwei and Tian, Wei and Chen, Dafang and Liu, Yajun and Wang, Lifang and Duan, Fangfang},
  journal={BMC musculoskeletal disorders},
  volume={19},
  pages={1--12},
  year={2018},
  publisher={Springer}
}

\clearpage

\section*{Appendix}\label{sec:appendix}

\begin{table}[!h]
\caption{Linear regression coefficients and 95\% confidence intervals (CIs) quantifying the association between cervical and thoracic degenerative conditions with spine-age gap. Separate regression models were fit for cervical and thoracic degenerate conditions. In all cases models controlled for biological sex.}
\centering
\begin{tabular}{ll|c|c}
\hline
\multicolumn{2}{c|}{Condition} & n & Effect (95\% CI) \\
\hline
\multirow{14}{*}{\shortstack{Cervical degenerative \\ spinal conditions}} 
& Mild disc bulge (No.$>$2) & 459 & 0.64 (0.19, 1.10)$^*$\\
& Moderate disc bulge (No.$>$1) & 32 & 2.26 (0.60, 3.92)$^*$ \\
& Severe disc bulge (No.$>$0) & 9 & 1.09 (-2.03, 4.21) \\
& Mild disc dessication (No.$>$1) & 243 & -0.23 (-0.85, 0.38) \\
& Moderate disc dessication (No.$>$0) & 210 & 0.44 (-0.23, 1.10) \\
& Severe disc dessication (No.$>$0) & 124 & -0.03 (-0.89, 0.84) \\
& Near complete disc dessication (No.$>$0) & 27 & -1.14 (-2.94, 0.67) \\
& Disc annular fissure (No.$>$0) & 79 & 0.18 (-0.88, 1.25) \\
& Vertebral endplate change (No.$>$0) & 208 & 0.06 (-0.61, 0.74) \\
& Mild disc osteophyte (No.$>$1) & 1805 & 1.12 (0.86, 1.38)$^*$ \\
& Moderate disc osteophyte (No.$>$0) & 276 & 1.09 (0.51, 1.67)$^*$ \\
& Mild disc protrusion (No.$>$1) & 271 & 0.43 (-0.15, 1.01) \\
& Moderate disc protrusion (No.$>$0) & 64 & 1.00 (-0.17, 2.18) \\
& Mild disc extrusion (No.$>$0) & 12 & -1.44 (-4.15, 1.26) \\
\hline
\multirow{10}{*}{\shortstack{Thoracic degenerative \\ spinal conditions}} 
& Mild disc bulge (No.$>$2) & 136 & 0.75 (-0.07, 1.58)\\
& Mild disc dessication (No.$>$1) & 96 & -0.04 (-1.03, 0.94) \\
& Moderate disc dessication (No.$>$0) & 43 & -1.01 (-2.44, 0.43) \\
& Severe disc dessication (No.$>$0) & 15 & -3.42 (-5.87, -0.96)$^*$ \\
& Disc annular fissure (No.$>$0) & 24 & 0.63 (-1.30, 2.56) \\
& Vertebral endplate change (No.$>$0) & 124 & 0.78 (-0.09, 1.65) \\
& Mild disc osteophyte (No.$>$1) & 30 & 1.13 (-0.59, 2.85) \\
& Mild disc protrusion (No.$>$1) & 177 & 0.67 (-0.04, 1.39) \\
& Moderate disc protrusion (No.$>$0) & 26 & 1.82 (-0.03, 3.66) \\
& Mild disc extrusion (No.$>$0) & 24 & -1.72 (-3.64, 0.20) \\
\hline
\multicolumn{4}{l}{\footnotesize{$^*$ statistically significant effects.}}
    \end{tabular}
\end{table}

\end{document}